\documentclass{article}
\usepackage{graphicx}
 \usepackage{amsmath}
\PassOptionsToPackage{numbers,compress}{natbib}
 \usepackage[preprint]{neurips_2026}

\usepackage[utf8]{inputenc} 
\usepackage[T1]{fontenc}    
\usepackage{hyperref}       
\usepackage{url}            
\usepackage{booktabs}       
\usepackage{amsfonts}       
\usepackage{nicefrac}       
\usepackage{microtype}      
\usepackage{xcolor}         

\title{When Decodability Is Not Enough: Logical Validity Representations, Behavioral Dissociation, and Causal Tests in Language Models}

\author{Smitha Muthya Sudheendra\\
  University of Minnesota, Twin Cities\\
  \texttt{muthy009@umn.edu} \\
  \And
  Jaideep Srivastava \\
  University of Minnesota, Twin Cities \\
  \texttt{srivasta@umn.edu} \\
}

\begin{document}

\maketitle

\begin{abstract}
Large language models can look capable of logical reasoning, but correct or incorrect answers alone tell us little about what the model represents internally. We study logical verification in five open-weight transformer models using matched valid--invalid premise--claim pairs that vary across inference families, semantic domains, templates, and difficulty levels. Despite near-chance behavioral performance, logical validity is often almost perfectly decodable from hidden states and remains strongly decodable under held-out templates, domains, and inference families. Validity also remains highly decodable on behaviorally incorrect examples in the conditions where correctness-conditioned evaluation is well defined. At the same time, exhaustive leave-one-out tests reveal clear limits to this generalization, and interventions along probe-derived validity directions have only weak, nonspecific effects compared with random controls. Our results suggest that representing validity, expressing it in behavior, and using it causally are distinct. Validity related information can be strongly decodable from a model's hidden states without being reliably expressed in its output.

\end{abstract}

\section{Introduction}
\label{sec:introduction}

Large language models are often evaluated on tasks described as requiring reasoning, but behavioral success alone tells us little about the internal computations behind an answer. A correct prediction may reflect the relevant logical relation, but it may also arise from lexical or semantic regularities. Likewise, an incorrect prediction does not necessarily mean that the information needed for the correct decision is absent from the model's hidden states. Distinguishing what a model outputs from what it represents, and how those
representations are used, is therefore central to interpretability.

We study this distinction through \emph{logical verification}: given a set of premises and a candidate claim, the model must determine whether the claim follows. This setting is useful because validity is defined by the relation between premises and claim, and gold labels can be assigned deterministically. It therefore supports controlled valid--invalid contrasts while varying surface form, semantic content, and inference structure. 

We separate three questions: whether the model expresses the correct validity judgment, whether validity-related information is linearly accessible in its hidden states, and whether the corresponding probe-derived direction causally influences the final decision. To study them, we construct 800 examples organized into 400 matched valid--invalid pairs spanning five inference families, five semantic domains, and three difficulty levels. We evaluate five open-weight transformers using behavioral likelihoods and layer-wise linear probes, test generalization to held-out templates, domains, and inference families, and use
matched-pair, correctness-conditioned, lexical, metadata, and shuffled-label controls. On three models, we additionally intervene along the probe-derived validity direction and compare its effect with norm-matched random directions.

Behavioral verification remains at or near chance across all five models, often with strong answer-label preferences. Hidden states tell a different story: validity is almost perfectly linearly decodable in-distribution and remains strongly decodable under held-out templates and many domain and inference-family shifts. Exhaustive leave-one-out evaluation nevertheless reveals clear exceptions, including a recurring failure to transfer to unseen syllogistic reasoning. Generalization is therefore broad but not uniform.

The dissociation also persists on behavioral errors. Where correctness-conditioned evaluation is well defined, validity remains highly decodable on incorrectly answered examples. Across all primary splits, probes also preserve the ordering between valid and invalid members of matched pairs, even when global calibration shifts. Behavioral errors therefore do not generally coincide with the absence of linearly accessible validity information. High decodability, however, does not imply simple causal control. Interventions
along the probe-derived direction produce only small and inconsistent changes in output margins, comparable to or smaller than random orthogonal controls. Thus, the linear direction that separates valid from invalid examples is not itself a strong control variable for verification behavior. Taken together, the results distinguish
\[
\text{behavioral}\ \text{expression}
\;\neq\;
\text{representational}\ \text{accessibility}
\;\neq\;
\text{causal}\ \text{use}.
\]
We study this separation in a controlled logical-verification setting, testing how well validity decodability generalizes and whether it persists when the model answers  incorrectly. More broadly, the results argue for treating behavioral performance, decodability, and causal influence as distinct forms of evidence when studying reasoning in language models.

\section{Related Work}

\paragraph{Reasoning and logical evaluation.}
Behavioral benchmarks such as BIG-Bench, HELM, and BIG-Bench Hard have documented substantial progress in language-model reasoning while also revealing strong sensitivity to task formulation and prompting 
\cite{srivastava2023beyond,liang2023holistic,suzgun2023challenging}. More targeted benchmarks, including FOLIO, PrOntoQA-OOD, LogicBench, and Multi-LogiEval, evaluate deductive reasoning and systematic generalization across logical structures and distribution shifts \citep{han2024folio,saparov2023testing,parmar2024logicbench,patel2024multi}. Prompting methods such as chain-of-thought can improve observable reasoning behavior \citep{wei2022chain,kojima2022large}, but generated explanations need not faithfully reflect the computation that produced an answer \citep{turpin2023language}. Our focus is therefore narrower: we study verification as a controlled relation between premises and a candidate claim, and distinguish output behavior from internal representation.

\paragraph{Internal representations of truth and validity.}
Prior work shows that high-level variables such as factual truth can be recovered from language-model activations \citep{burns2023discovering,marks2023geometry}, while representation engineering studies whether such directions can also be manipulated \citep{zou2023representation}. Logical validity is related but distinct: it depends on the relation between premises and a conclusion rather than on the truth of the conclusion alone. Our matched-pair construction is designed to isolate this relational property. \citet{bertolazzi-etal-2026-language} show that logical validity and semantic plausibility are linearly represented in syllogistic reasoning and can be causally influenced through activation steering. We instead ask how broadly validity decodability generalizes across surface form, semantic domain, and inference structure, and whether validity remains accessible when the model's behavioral judgment is incorrect.

\paragraph{Probing and causal interpretation.}
High probe accuracy does not imply that the decoded feature is used by the model. Probe performance can reflect nuisance variables or task-format artifacts, and recent work shows that apparently strong reasoning-related separability can disappear after such confounds are controlled \citep{sahoo-etal-2026-linear}. We therefore use matched valid--invalid pairs, held-out template/domain/family evaluations, shuffled-label controls, and metadata-only baselines. Throughout, we interpret probe AUROC as evidence of linear accessibility, not as evidence of an abstract reasoning mechanism. Mechanistic interpretability further motivates testing whether decoded structure is behaviorally consequential. Prior work has recovered internal state variables through probing and intervention in settings such as Othello-GPT and multi-step reasoning \citep{li2023emergent,nanda2023emergent,hou-etal-2023-towards}. We similarly complement layer-wise probing with activation interventions along probe-derived validity directions and compare them with norm-matched random controls.

Rather than asking only whether logical validity can be decoded from hidden states, we ask how meaningful that representation is. Does it generalize beyond the conditions used to train the probe? Is it still present when the model gives the wrong answer? And does intervening on the corresponding direction actually change the model's behavior? These questions let us separate what is \emph{represented} internally from what is \emph{expressed} in the output and what is \emph{used} causally in the model's decision.

\section{Controlled Verification Dataset}
\label{sec:dataset}

We construct a controlled dataset to study how language models represent logical validity. Rather than aiming for broad benchmark coverage, the dataset is designed to isolate verification from confounds such as factual recall, search, and explanation generation. Each example consists of a set of premises and a candidate claim, and the model must decide whether the claim follows from the premises. Because validity depends on the relation between the premises and the claim, this setting allows us to study relational information rather than claim plausibility alone.

\subsection{Task and Dataset Structure}

Each example is represented as
\[
x_i=(P_i,c_i,y_i,f_i,d_i,t_i,\delta_i),
\]
where $P_i$ is the set of premises, $c_i$ the candidate claim, and $y_i\in\{0,1\}$ the validity label. The remaining variables denote inference family, semantic domain, template family, and difficulty level.

Models choose between \texttt{VALID} and \texttt{INVALID}. The dataset contains five inference families: syllogism, transitivity, set inclusion, causal chain, and permission logic, instantiated across five domains: nonce, spatial, social, biological, and legal-policy. Using the same inference structures across different domains lets us test whether validity representations depend on particular semantic content.

We also define three difficulty levels:
\begin{itemize}
    \item \textbf{Difficulty 1:} direct one-step verification;
    \item \textbf{Difficulty 2:} two-step inference chains;
    \item \textbf{Difficulty 3:} two-step chains with an irrelevant distractor premise.
\end{itemize}

These levels vary the amount of relational integration required while leaving the output task unchanged.

\subsection{Matched Valid--Invalid Pairs}

The dataset is organized into matched valid--invalid pairs. Within each pair, the premises, inference family, domain, template family, and difficulty are held fixed; only the candidate claim changes.

For a matched pair $(x_i^{+},x_i^{-})$,
\[ P_i^{+}=P_i^{-},\quad f_i^{+}=f_i^{-},\quad d_i^{+}=d_i^{-},\quad t_i^{+}=t_i^{-},\quad \delta_i^{+}=\delta_i^{-}, \]
while
\[ y_i^{+}=1,\qquad y_i^{-}=0. \]

This construction reduces label-correlated differences in the surrounding context and gives us a direct within-context test: does the model representation distinguish a valid claim from an invalid one when everything else is held constant? Matched pairs are also kept together during splitting and bootstrap resampling.

\subsection{Dataset Composition}
The final dataset contains 800 examples, equally divided between valid and invalid inferences. Each of the five inference families contributes 160 examples, and each of the five semantic domains contributes 160 examples. The three difficulty levels contain 200 one-step examples, 400 two-step examples, and 200 two-step examples with distractor premises. The strict template split contains 600 training examples and 200 held-out examples.

\begin{table}[t]
\centering
\small
\caption{Composition of the controlled verification dataset.}
\label{tab:dataset_composition}
\setlength{\tabcolsep}{8pt}
\renewcommand{\arraystretch}{1.08}
\begin{tabular}{lr}
\toprule
\textbf{Property} & \textbf{Count} \\
\midrule
Total examples & 800 \\
Valid / Invalid & 400 / 400 \\
Inference families & 5 \\
Semantic domains & 5 \\
Examples per family & 160 \\
Examples per domain & 160 \\
Difficulty 1 / 2 / 3 & 200 / 400 / 200 \\
Template train / test & 600 / 200 \\
\bottomrule
\end{tabular}
\end{table}

\subsection{Generalization Splits}
\label{sec:generalization_splits}

Random held-out performance shows whether validity is decodable in-distribution, but it does not tell us whether the probe depends on familiar wording, semantic content, or inference-specific structure. We therefore evaluate four complementary train--test conditions.

The \emph{random split} serves as the in-distribution reference, with training and test examples drawn from the same overall mixture. In the \emph{template-held-out split}, complete template families are excluded from training to test whether decoding transfers to unseen surface forms. The \emph{domain-held-out split} evaluates the probe on a semantic domain that is absent during training, while the \emph{inference-family-held-out split} withholds an entire reasoning family and tests transfer to a new inference structure.

These conditions are not intended to form a strict hierarchy of difficulty. Rather, they probe different kinds of generalization: from ordinary in-distribution prediction to changes in surface form, semantic content, and inference structure.

\subsection{Controls and Scope}

All examples are generated deterministically from known inference rules, so gold labels are assigned by construction rather than by a language model. We retain metadata describing inference family, domain, template family, difficulty, distractor status, and split membership for later subgroup and control analyses.

The dataset is intentionally diagnostic rather than comprehensive. Its purpose is to distinguish several questions: whether models express validity in their outputs, whether validity-related information is accessible in hidden states, whether it generalizes beyond the probe-training distribution, and whether it remains accessible when the final answer is wrong.

\section{Experimental Setup}
\label{sec:setup}

We evaluate logical verification at three levels: output behavior, linear accessibility in hidden states, and the causal effect of the probe-derived validity direction. Full implementation and statistical details are provided in Appendix~\ref{app:implementation} and Appendix~\ref{app:probing}.

\subsection{Models and Behavioral Evaluation}
\label{sec:behavior}

We evaluate Pythia-1.4B, Pythia-2.8B, SmolLM3-3B, Llama-3.2-3B (hereafter, Llama-3.2), and Mistral-7B. Causal interventions are run on Pythia-2.8B, Llama-3.2, and Mistral-7B. All models are evaluated on the same 800 examples and split definitions. Behavior is measured from the likelihood assigned to \texttt{VALID} and \texttt{INVALID}. For label scores $s_V(x)$ and $s_I(x)$, we predict
\[
\hat y(x)=\mathbb{I}[s_V(x)>s_I(x)],
\qquad
M(x)=s_V(x)-s_I(x),
\]
where $M(x)$ is the continuous output margin. We report accuracy, the fraction of predictions assigned \texttt{VALID}, and margin AUROC. Answer-label tokenization is audited separately for each model; details are given in Appendix~\ref{app:implementation}.

\subsection{Hidden-State Probing}
\label{sec:probing}

For every transformer block, we extract the final prompt-token hidden state and fit an $\ell_2$-regularized logistic-regression probe to predict gold validity. Feature standardization, hyperparameter selection, and layer selection use only the probe-training partition, with matched pairs kept together during validation. The reported layer is \[
\ell^*=\arg\max_\ell
\operatorname{AUROC}^{\mathrm{val}}_\ell,
\]
after which the selected probe is evaluated once on the held-out test set.

We evaluate probes under random, template-held-out, domain-held-out, and inference-family-held-out splits, and additionally run exhaustive leave-one-domain-out and leave-one-family-out analyses. Matched-pair accuracy and correctness-conditioned AUROC are used to test whether validity remains accessible within controlled pairs and on behaviorally incorrect examples. Full probe and resampling details are given in
Appendix~\ref{app:probing}.

\subsection{Controls}
\label{sec:controls}

We compare hidden-state probes with TF--IDF classifiers trained on the full prompt, claim only, and premises only. We also fit metadata-only classifiers and repeat probing with 200 shuffled-label permutations. These controls test whether probe performance can be explained by lexical regularities, construction metadata, or arbitrary linear separability. Difficulty and within-Pythia scaling are treated as secondary analyses and reported in Appendix~\ref{app:secondary}.

\subsection{Causal Intervention}
\label{sec:causal}

To test whether the probe-derived direction is behaviorally consequential, we intervene on the final prompt-token state at the selected layer. The normalized probe direction is expressed in raw activation coordinates and scaled by the training-set standard deviation of projection onto that direction. We apply interventions at
\[
\alpha\in\{-4,-2,-1,0,1,2,4\},
\]
and measure the resulting change in output margin
\[
\Delta M_\alpha(x)=M_\alpha(x)-M_0(x).
\]

The learned direction is compared with five norm-matched random orthogonal directions. We additionally perform matched-projection patching between valid and invalid members of each pair. These experiments test the narrow hypothesis that the particular linear direction identified by the probe is sufficient to influence verification behavior. Full intervention construction and pair-bootstrap procedures are given in Appendix~\ref{app:causal}.

\section{Results}
\label{sec:results}

Across the five models, behavioral verification and internal validity decodability diverge sharply. Output-level performance remains at or near chance, while validity is strongly linearly decodable from hidden states. This decodability transfers well across unseen templates and many semantic domains and inference families, although exhaustive leave-one-out evaluation reveals systematic exceptions. Finally, the probe-derived validity direction has only weak and non-specific effects on model outputs.

\subsection{Behavioral Verification Is Unreliable}
\label{sec:behavior_results}

Behavioral accuracy is close to chance for every model (Table~\ref{tab:behavior_results}), but the response patterns differ substantially. Pythia-1.4B predicts \texttt{INVALID} for every example and Pythia-2.8B does so for nearly all examples. SmolLM3-3B and Mistral-7B show the opposite preference, predicting \texttt{VALID} throughout. Llama-3.2 uses both labels but reaches only $0.470$ accuracy. Margin AUROC is also near chance for Pythia and Llama. SmolLM3 and Mistral retain modest ranking information, but this is not reflected in their binary predictions. Thus, similar chance-level accuracies mask very different model-specific answer-label biases.

\begin{table*}[t]
\caption{Behavioral verification performance. Confidence intervals use
pair-level bootstrap resampling.}
\label{tab:behavior_results}
\centering
\small
\setlength{\tabcolsep}{7pt}
\renewcommand{\arraystretch}{1.08}
\begin{tabular}{lccc}
\toprule
\textbf{Model}
& \textbf{Accuracy [95\% CI]}
& \textbf{Pred. VALID}
& \textbf{Margin AUROC [95\% CI]} \\
\midrule
Pythia-1.4B
& 0.500 [0.500, 0.500]
& 0.000
& 0.495 [0.488, 0.503] \\

Pythia-2.8B
& 0.500 [0.496, 0.504]
& 0.013
& 0.496 [0.487, 0.505] \\

SmolLM3-3B
& 0.500 [0.500, 0.500]
& 1.000
& 0.561 [0.551, 0.572] \\

Llama-3.2-3B
& 0.470 [0.455, 0.483]
& 0.555
& 0.458 [0.450, 0.464] \\

Mistral-7B
& 0.500 [0.500, 0.500]
& 1.000
& 0.576 [0.564, 0.588] \\
\bottomrule
\end{tabular}
\end{table*}

\subsection{Validity Is Strongly Decodable from Hidden States}
\label{sec:probe_results}

The hidden-state results look very different from the behavioral results. Random-split AUROC is essentially perfect for every model
(Table~\ref{tab:probe_results}). Validity-related information is also linearly accessible very early in the network: at the first transformer block, random-split AUROC ranges from 0.935 to 0.995 across models.

Decodability remains high under unseen templates, with AUROC between $0.963$ and $0.999$. Most domain- and family-held-out evaluations are also strong, although the exceptions become clearer in the exhaustive analysis below. The selected layer varies substantially across architectures, from normalized depth $0.214$ for Llama-3.2 to $0.556$ for SmolLM3-3B.

\begin{table*}[t]
\caption{Selected-layer validity-probe AUROC. $\ell^*/L$ gives the selected
random-split layer and normalized depth. Confidence intervals use pair-level
bootstrap resampling.}
\label{tab:probe_results}
\centering
\scriptsize
\setlength{\tabcolsep}{5pt}
\renewcommand{\arraystretch}{1.08}
\begin{tabular}{lccccc}
\toprule
\textbf{Model}
& $\boldsymbol{\ell^*/L}$
& \textbf{Random}
& \textbf{Template}
& \textbf{Domain}
& \textbf{Family} \\
\midrule
Pythia-1.4B
& 10/24 (0.417)
& 1.000 [1.000, 1.000]
& 0.999 [0.996, 1.000]
& 0.968 [0.949, 0.987]
& 0.950 [0.921, 0.973] \\

Pythia-2.8B
& 10/32 (0.313)
& 1.000 [1.000, 1.000]
& 0.963 [0.936, 0.986]
& 0.989 [0.976, 0.998]
& 0.976 [0.959, 0.990] \\

SmolLM3-3B
& 20/36 (0.556)
& 1.000 [0.999, 1.000]
& 0.988 [0.978, 0.996]
& 0.982 [0.966, 0.994]
& 0.938 [0.911, 0.965] \\

Llama-3.2-3B
& 6/28 (0.214)
& 1.000 [1.000, 1.000]
& 0.980 [0.962, 0.995]
& 0.991 [0.975, 1.000]
& 0.992 [0.983, 0.998] \\

Mistral-7B
& 12/32 (0.375)
& 1.000 [1.000, 1.000]
& 0.987 [0.977, 0.995]
& 0.771 [0.740, 0.814]
& 0.995 [0.987, 0.999] \\
\bottomrule
\end{tabular}
\end{table*}

\begin{figure*}[t]
\centering

\includegraphics[width=0.49\linewidth]{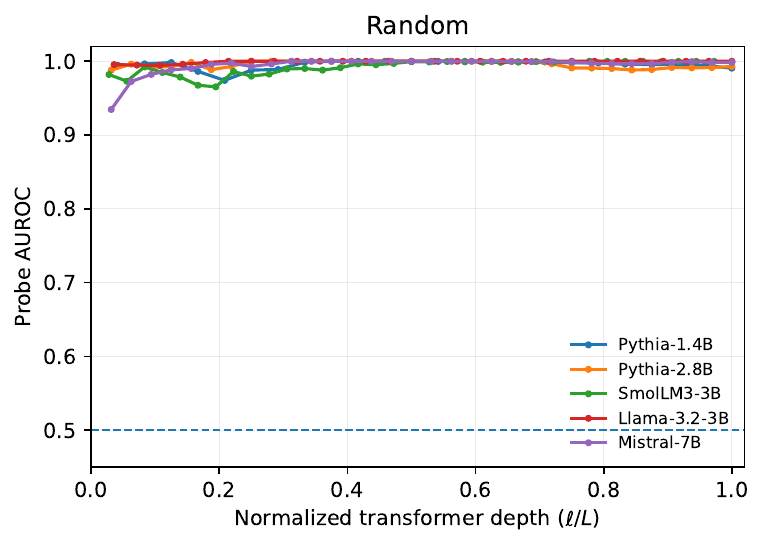}
\hfill
\includegraphics[width=0.49\linewidth]{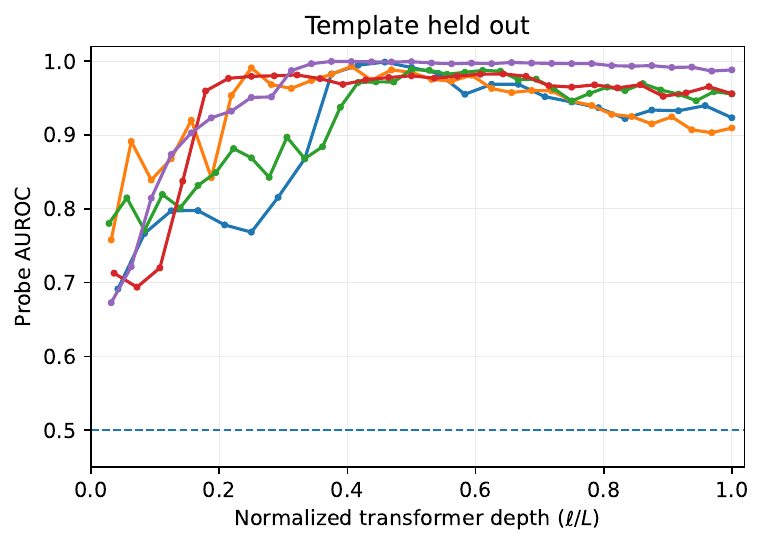}

\vspace{0.4em}

\includegraphics[width=0.49\linewidth]{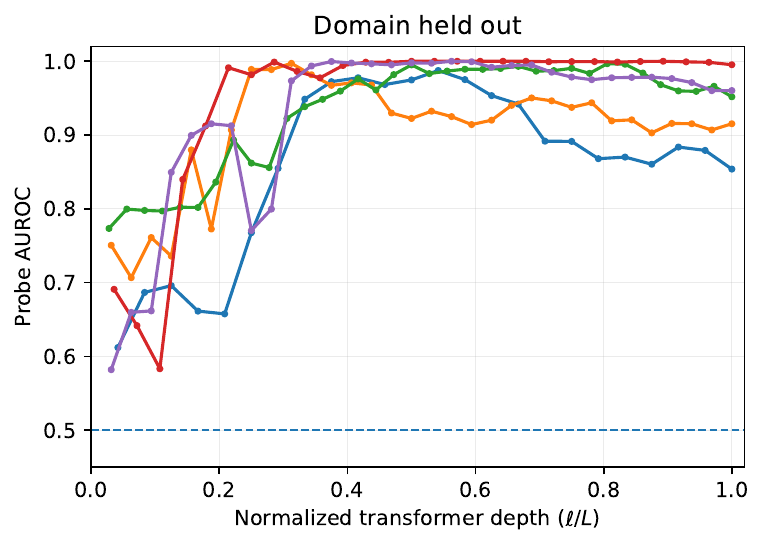}
\hfill
\includegraphics[width=0.49\linewidth]{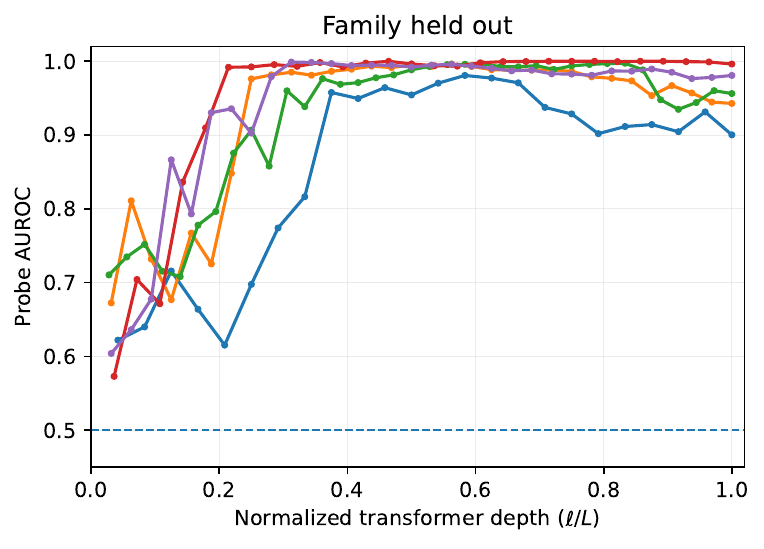}

\caption{Layer-wise validity-probe AUROC across normalized transformer depth.
Top left: random split; top right: template held out; bottom left: domain held
out; bottom right: inference family held out. Validity remains broadly
decodable under distribution shift, although the trajectories differ across
models and conditions.}
\label{fig:layerwise_probe}
\end{figure*}

Figure~\ref{fig:layerwise_probe} shows the full layer-wise trajectories. In-distribution decodability stays close to ceiling across most of the network. Under distribution shift, the early layers are more variable, but most models reach high AUROC by the middle layers.

\subsection{OOD Generalization Is Broad but Not Uniform}
\label{sec:ood_results}

\begin{figure*}[t]
\centering

\includegraphics[width=0.49\linewidth]{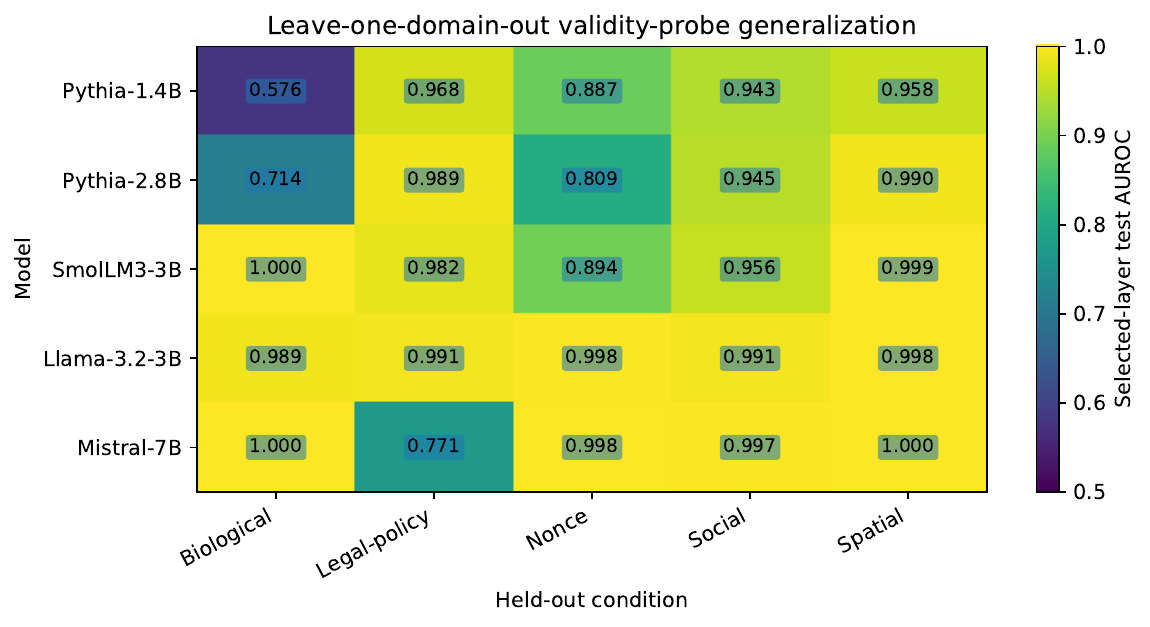}
\hfill
\includegraphics[width=0.49\linewidth]{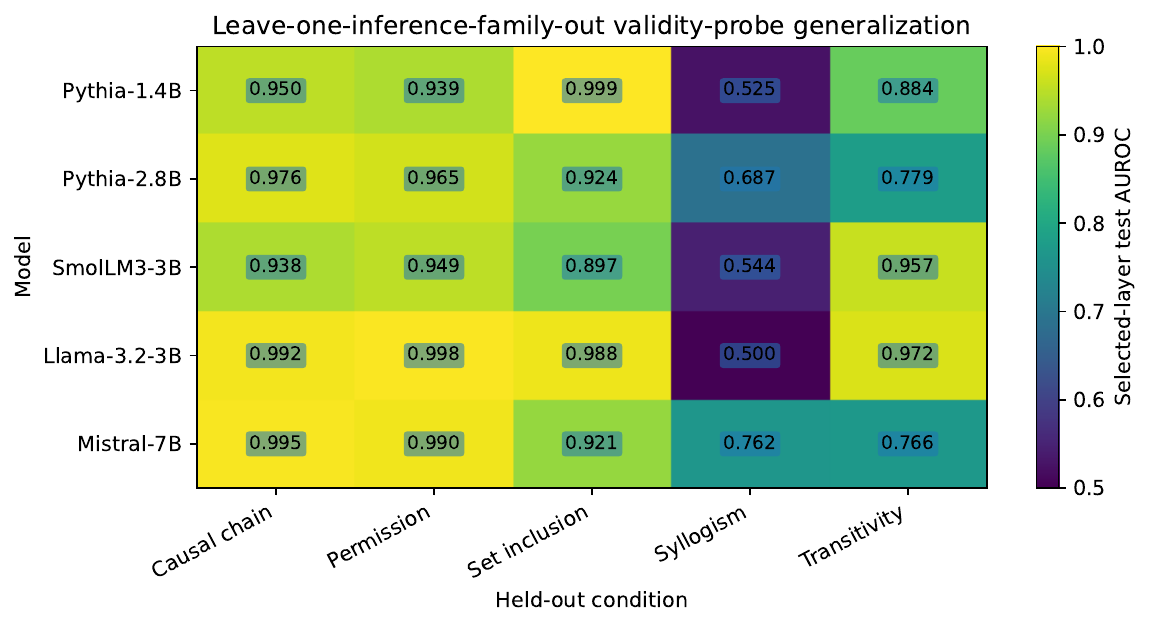}

\caption{Exhaustive leave-one-out validity-probe generalization.
Left: each semantic domain is held out in turn. Right: each inference family
is held out. Domain failures are largely model-specific, whereas syllogistic
reasoning is consistently the weakest unseen inference family.}
\label{fig:loo_heatmaps}
\end{figure*}

The predefined domain and family splits show strong transfer for most models, but exhaustive leave-one-out evaluation reveals substantial heterogeneity (Figure~\ref{fig:loo_heatmaps}). Domain transfer is particularly stable for Llama-3.2, which remains at or above $0.989$ for every held-out domain. Other models show more localized failures. Pythia-1.4B and Pythia-2.8B fall to $0.576$ and $0.714$ when the
biological domain is withheld, while Mistral-7B falls to $0.771$ on legal-policy.  The family results show a more consistent weakness. Syllogism is the weakest held-out inference family for every model: AUROC falls to $0.525$ for Pythia-1.4B, $0.687$ for Pythia-2.8B, $0.544$ for SmolLM3-3B, $0.500$ for Llama-3.2, and $0.762$ for Mistral-7B. Several other families transfer close to perfectly. These results support substantial shared validity-related structure across conditions, but not a single representation that transfers uniformly across all semantic domains and inference families. Full leave-one-out values are reported in Appendix~\ref{app:generalization}.

\subsection{Validity Remains Decodable When Behavior Fails}
\label{sec:dissociation_results}

Matched-pair evaluation gives a strong within-context result. Across all five models and all four primary splits, the selected probes achieve pairwise accuracy of $1.000$. This ordering remains intact even when global calibration degrades. Mistral-7B, for example, reaches only $0.771$ global AUROC under domain holdout while still ranking the valid member above the invalid member of every matched pair. Correctness-conditioned analysis gives a complementary view. Llama-3.2 provides the cleanest comparison because both gold classes are represented in the correct and incorrect subsets. AUROC on incorrectly answered examples is $1.000$, $0.995$, $1.000$, and $0.994$ for the random, template, domain, and family splits, respectively. Pythia-2.8B shows the same qualitative pattern where the comparison is defined; under domain holdout, AUROC is $0.965$ on correct examples and $1.000$ on incorrect examples.

For models with nearly deterministic answer-label preferences, some correctness-conditioned AUROCs are undefined because the resulting subset contains only one gold class. We therefore make the narrower claim that, where this comparison is statistically well defined, behavioral errors do not generally coincide with the absence of linearly accessible validity information. Full results are given in Appendix~\ref{app:dissociation}.

\subsection{Controls Qualify the In-Distribution Result}
\label{sec:control_results}

The random split contains substantial lexical predictability. Full-prompt TF--IDF reaches $0.970$ AUROC and claim-only TF--IDF reaches $0.965$, so near-perfect in-distribution probe performance alone is not sufficient evidence for a general validity representation. The lexical baselines weaken more under distribution shift. Full-prompt TF--IDF falls to $0.855$ under template holdout and $0.814$ under domain
holdout, while claim-only performance falls to $0.682$ and $0.767$, respectively. Hidden-state probes are generally stronger under these shifts, although Mistral's domain-held-out result ($0.771$) is an exception. Premises-only classifiers remain at chance, and metadata-only classifiers remain between approximately $0.49$ and $0.51$. Shuffled-label probes also remain near chance: null means range from $0.496$ to $0.499$, with standard deviations of approximately $0.031$--$0.039$. For every model, the observed random-split result exceeds all 200 permutations ($p=1/201\approx0.005$). These controls indicate that surface regularities contribute substantially to in-distribution separability, but do not fully account for the broader hidden-state generalization. Full control results are reported in Appendix~\ref{app:controls}.

\paragraph{Scaling and difficulty.} Within the Pythia family, increasing scale from 1.4B to 2.8B improves domain- and family-held-out performance but reduces template-held-out AUROC from $0.999$ to $0.963$. Behavioral accuracy also shows no consistent decline across the three nominal difficulty levels. We therefore find neither a uniform within-family scaling effect nor a monotonic behavioral difficulty gradient. Full results are reported in Appendix~\ref{app:secondary}.

\subsection{Strong Decodability Does Not Imply Causal Control}
\label{sec:causal_results}

We finally test the narrower causal hypothesis that the linear direction identified by the probe is itself sufficient to influence verification behavior at the selected intervention site. At $\alpha=+4$, interventions along the probe-derived direction change the \texttt{VALID}--\texttt{INVALID} margin by only $-0.0037$ for Pythia-2.8B, $-0.0022$ for Llama-3.2, and $+0.0023$ for Mistral-7B (Table~\ref{tab:causal_results}). The sign is not consistent across models, and random orthogonal perturbations produce effects of comparable or greater magnitude.

\begin{table*}[t]
\caption{Causal intervention results at $\alpha=+4$. $\Delta M$ is the mean
change in the \texttt{VALID}--\texttt{INVALID} output margin. Random columns
summarize five norm-matched orthogonal directions.}
\label{tab:causal_results}
\centering
\small
\setlength{\tabcolsep}{6pt}
\renewcommand{\arraystretch}{1.08}
\begin{tabular}{lcccc}
\toprule
\textbf{Model}
& $\boldsymbol{\Delta M}$ \textbf{ [95\% CI]}
& \textbf{Prediction flips}
& \textbf{Mean $|\Delta M|$ random}
& \textbf{Max $|\Delta M|$ random} \\
\midrule
Pythia-2.8B
& $-0.0037$ [$-0.0076$, $0.0004$]
& 0.0\%
& 0.0077
& 0.0129 \\

Llama-3.2-3B
& $-0.0022$ [$-0.0031$, $-0.0013$]
& 0.0\%
& 0.0072
& 0.0097 \\

Mistral-7B
& $+0.0023$ [$+0.0011$, $+0.0034$]
& 0.0\%
& 0.0058
& 0.0103 \\
\bottomrule
\end{tabular}
\end{table*}
No model changes its binary prediction under the $\alpha=+4$ intervention. At $\alpha=-4$, only one of 160 Llama examples changes prediction. Matched-projection patching gives the same qualitative result, with margin changes remaining close to zero. These results do not show that validity-related information is causally irrelevant. They show that the particular linear probe direction, at the
tested intervention site, is not a strong or validity-specific control variable for the final decision. Full intervention details and the complete strength sweep are reported in Appendix~\ref{app:causal}.
\section{Discussion}
Validity can still be easy to decode from the hidden states even when the model gets the verification decision wrong. This is most apparent in the matched-pair and correctness-conditioned analyses, where decodability remains high on examples the model answers incorrectly. At the same time, the representation does not generalize uniformly. Leave-one-out evaluation reveals model-specific domain failures and a recurring weakness when syllogistic reasoning is excluded from probe training. Validity-related structure therefore appears to be partly shared across reasoning settings, but not fully invariant across them. The intervention results further limit what can be concluded from high probe accuracy. Manipulating the probe-derived direction produces only small and inconsistent behavioral changes, often no larger than random-direction controls. Linear accessibility should therefore be distinguished from both behavioral expression and causal control.
\section{Limitations}

Our dataset is deliberately controlled and synthetic. This makes matched comparisons and distribution shifts possible, but limits how directly the findings generalize to naturalistic reasoning, longer contexts, or more open-ended tasks. The five inference families and semantic domains also cover only a small part of the space of logical reasoning.

Our analysis focuses on linear probes at the final prompt-token representation. Validity may also be encoded nonlinearly, distributed across token positions, or represented differently at other computational sites. Likewise, the weak steering effects only rule out a relatively simple causal interpretation of the probe direction; they do not show that validity-related information is causally irrelevant to the model's computation.

Finally, all models are relatively small open-weight transformers evaluated under quantization. Replication with larger models and a broader range of architectures would help establish how general these patterns are.

\section{Conclusion}

We studied logical verification at three levels: behavioral expression, representational accessibility, and causal use.
For models and conditions where correctness-conditioned evaluation is well defined, validity remains strongly decodable even on incorrectly answered examples. However, generalization is not universal, and probe-derived validity directions have little specific causal effect on output behavior. These findings show why high probe accuracy should be interpreted carefully. A feature can be readily recoverable from a model's representations without being reliably expressed in its output or corresponding to a simple causal mechanism for the final decision.

\bibliographystyle{plainnat}
\bibliography{custom}


\appendix

\appendix

\section{Dataset Construction}
\label{app:dataset}

\subsection{Task Representation}

The dataset contains 800 examples organized into 400 matched valid--invalid pairs. Each example is represented as 
\[
x_i=(P_i,c_i,y_i,f_i,d_i,t_i,\delta_i),
\]
where $P_i$ is the premise set, $c_i$ is the candidate claim, $y_i\in\{0,1\}$ is the validity label, $f_i$ denotes inference family, $d_i$ semantic domain, $t_i$ template family, and $\delta_i$ difficulty. We instantiate five inference families: syllogism, transitivity, set inclusion, causal chain, and permission logic. Each family appears across five semantic domains: nonce, spatial, social, biological, and legal-policy.

\subsection{Matched Valid--Invalid Construction}

For each pair, the premise context and construction variables are held fixed
while the candidate claim changes. For a matched pair
$(x_i^+,x_i^-)$,
\[
P_i^+=P_i^-,
\qquad
f_i^+=f_i^-,
\qquad
d_i^+=d_i^-,
\qquad
t_i^+=t_i^-,
\qquad
\delta_i^+=\delta_i^-,
\]
while
\[
y_i^+=1,
\qquad
y_i^-=0.
\]

Matched pairs are kept together during data splitting, probe validation,
and pair-level bootstrap resampling.

\subsection{Difficulty Levels}

We use three controlled difficulty levels:
\begin{itemize}
    \item \textbf{Difficulty 1:} direct one-step verification;
    \item \textbf{Difficulty 2:} two-step inference chains;
    \item \textbf{Difficulty 3:} two-step chains with an irrelevant
    distractor premise.
\end{itemize}

These labels describe the construction of an example rather than assuming that behavioral difficulty must increase monotonically across levels. 
\begin{table}[t]
\caption{Composition of the controlled verification dataset.}
\label{tab:appendix_dataset}
\centering
\small
\begin{tabular}{lr}
\toprule
\textbf{Property} & \textbf{Count} \\
\midrule
Total examples & 800 \\
Matched pairs & 400 \\
Valid / Invalid & 400 / 400 \\
Inference families & 5 \\
Semantic domains & 5 \\
Examples per family & 160 \\
Examples per domain & 160 \\
Difficulty 1 / 2 / 3 & 200 / 400 / 200 \\
Distractor examples & 200 \\
\bottomrule
\end{tabular}
\end{table}

\subsection{Train--Test Splits}

We evaluate four complementary probe splits. The random split is the in-distribution reference. The template-held-out split excludes complete template families from probe training, the domain-held-out split excludes a semantic domain, and the inference-family-held-out split excludes an entire reasoning family.

\begin{table}[t]
\caption{Primary probe train--test splits.}
\label{tab:appendix_splits}
\centering
\small
\begin{tabular}{lrr}
\toprule
\textbf{Split} & \textbf{Train} & \textbf{Test} \\
\midrule
Random & 640 & 160 \\
Template held out & 600 & 200 \\
Domain held out & 640 & 160 \\
Inference family held out & 640 & 160 \\
\bottomrule
\end{tabular}
\end{table}

All four splits remain label balanced. Complete template families are disjoint between training and test in the strict template condition. We additionally perform exhaustive leave-one-domain-out and leave-one-inference-family-out evaluations.

\section{Models and Implementation Details}
\label{app:implementation}


Layer position is reported using normalized transformer depth,
\[
d_\ell=\frac{\ell}{L},
\]
where $\ell$ is the block index and $L$ is the total number of blocks.

\begin{table*}[t]
\caption{Models used in the experiments.}
\label{tab:appendix_models}
\centering
\small
\setlength{\tabcolsep}{7pt}
\begin{tabular}{lclc}
\toprule
\textbf{Model} & \textbf{Parameters}
& \textbf{Hugging Face identifier} & \textbf{Layers} \\
\midrule
Pythia-1.4B
& 1.4B
& \texttt{EleutherAI/pythia-1.4b}
& 24 \\

Pythia-2.8B
& 2.8B
& \texttt{EleutherAI/pythia-2.8b}
& 32 \\

SmolLM3-3B
& 3B
& \texttt{HuggingFaceTB/SmolLM3-3B-Base}
& 36 \\

Llama-3.2-3B
& 3B
& \texttt{meta-llama/Llama-3.2-3B}
& 28 \\

Mistral-7B
& 7B
& \texttt{mistralai/Mistral-7B-v0.3}
& 32 \\
\bottomrule
\end{tabular}
\end{table*}

\subsection{Behavioral Label Scoring and Tokenization}

Behavior is scored from the conditional likelihood assigned to \texttt{VALID} and \texttt{INVALID}. Because these strings may tokenize differently across model families, we audit three candidate answer prefixes: the plain label, a leading-space label, and a leading-newline label. Prefix selection prioritizes equal token counts for the two answer labels, followed by the shortest available encoding.

For token sequences
$T_V=(v_1,\ldots,v_{K_V})$ and
$T_I=(u_1,\ldots,u_{K_I})$, we define
\[
s_V(x)=
\frac{1}{K_V}
\sum_{k=1}^{K_V}
\log p(v_k\mid x,v_{<k}),
\]
with $s_I(x)$ defined analogously. The behavioral margin is
\[
M(x)=s_V(x)-s_I(x).
\]

This audit is important because the strong answer-label preferences observed in Section~\ref{sec:behavior_results} should not be attributed simply to unequal answer-string length.

\section{Probe Training and Statistical Details}
\label{app:probing}

For each example $i$ and transformer block $\ell$, we extract the final prompt-token hidden state
\[
h_{i,\ell}\in\mathbb{R}^{m}.
\]

At each layer, we fit an $\ell_2$-regularized logistic-regression probe to predict gold validity. Features are standardized using statistics estimated only from the probe-training partition. The regularization coefficient is selected from
\[
C\in\{0.1,1,10\}
\]
using group-aware cross-validation with \texttt{pair\_id} as the grouping
variable.

Layer selection is also performed entirely within the training data:
\[
\ell^*
=
\arg\max_\ell
\operatorname{AUROC}^{\mathrm{val}}_\ell.
\]

After selecting $\ell^*$ and the regularization coefficient, the probe is refit on the available training data and evaluated once on the held-out test partition. Full layer-wise trajectories are shown in Figure~\ref{fig:layerwise_probe}.

\subsection{Pairwise Evaluation}

For each matched valid--invalid pair, we ask whether the probe assigns a larger validity score to the valid member:
\[
\operatorname{PairAcc}
=
\frac{1}{P}
\sum_{p=1}^{P}
\mathbb{I}
\left[q(i_V)>q(i_I)\right],
\]
where $q(\cdot)$ denotes the probe score.

\subsection{Uncertainty and Statistical Reporting}
\label{app:reporting}

Because examples occur in matched pairs, the pair is treated as the independent resampling unit. We compute 95\% confidence intervals from 2000 pair-level bootstrap resamples, sampling \texttt{pair\_id}s with replacement and recomputing the corresponding statistic. Probe performance is reported using AUROC. Behavioral evaluation reports accuracy, the fraction of predictions assigned \texttt{VALID}, and output-margin AUROC. Matched-pair analyses report pairwise accuracy. Correctness-conditioned AUROC is reported only when both gold classes are present in the corresponding subset; otherwise it is undefined and omitted. For shuffled-label controls, we report the mean and standard deviation of the null AUROC distribution over 200 permutations and compute 
\[
p=
\frac{
1+\#\{\text{permuted AUROC}\geq\text{observed AUROC}\}
}{
200+1
}.
\]

Unless otherwise stated, reported probe values refer to the layer selected using training-partition validation only.
\section{Full Generalization Results}
\label{app:generalization}

The primary domain-held-out split withholds legal-policy, while the primary inference-family-held-out split withholds causal chain. To test whether these results depend on that choice, we additionally hold out every semantic domain and inference family in turn.

\begin{table*}[t]
\caption{Leave-one-domain-out selected-layer validity-probe AUROC.}
\label{tab:domain_loo}
\centering
\small
\setlength{\tabcolsep}{6pt}
\begin{tabular}{lccccc}
\toprule
\textbf{Model}
& \textbf{Biological}
& \textbf{Legal-policy}
& \textbf{Nonce}
& \textbf{Social}
& \textbf{Spatial} \\
\midrule
Pythia-1.4B
& 0.576 & 0.968 & 0.887 & 0.943 & 0.958 \\

Pythia-2.8B
& 0.714 & 0.989 & 0.809 & 0.945 & 0.990 \\

SmolLM3-3B
& 1.000 & 0.982 & 0.894 & 0.956 & 0.999 \\

Llama-3.2-3B
& 0.989 & 0.991 & 0.998 & 0.991 & 0.998 \\

Mistral-7B
& 1.000 & 0.771 & 0.998 & 0.997 & 1.000 \\
\bottomrule
\end{tabular}
\end{table*}

\begin{table*}[t]
\caption{Leave-one-inference-family-out selected-layer validity-probe AUROC.}
\label{tab:family_loo}
\centering
\small
\setlength{\tabcolsep}{6pt}
\begin{tabular}{lccccc}
\toprule
\textbf{Model}
& \textbf{Causal chain}
& \textbf{Permission}
& \textbf{Set inclusion}
& \textbf{Syllogism}
& \textbf{Transitivity} \\
\midrule
Pythia-1.4B
& 0.950 & 0.939 & 0.999 & 0.525 & 0.884 \\

Pythia-2.8B
& 0.976 & 0.965 & 0.924 & 0.687 & 0.779 \\

SmolLM3-3B
& 0.938 & 0.949 & 0.897 & 0.544 & 0.957 \\

Llama-3.2-3B
& 0.992 & 0.998 & 0.988 & 0.500 & 0.972 \\

Mistral-7B
& 0.995 & 0.990 & 0.921 & 0.762 & 0.766 \\
\bottomrule
\end{tabular}
\end{table*}

The domain results show largely model-specific weaknesses. Llama-3.2 is the most stable across domains, whereas both Pythia models weaken when the biological domain is unseen and Mistral-7B weakens on legal-policy. The inference-family results reveal a more systematic pattern. Syllogism is the weakest unseen family for all five models. This heterogeneity motivates interpreting the main held-out results as evidence for substantial shared validity-related structure rather than a single representation that is fully invariant across reasoning families.

\section{Behavior--Representation Dissociation Details}
\label{app:dissociation}

Across all five models and all four primary evaluation conditions, the selected probes achieve matched-pair accuracy of $1.000$, with pair-level bootstrap confidence intervals of $[1.000,1.000]$. Correctness-conditioned evaluation asks whether validity remains decodable after separating examples according to whether the model's output-level
decision is correct. Llama-3.2 provides the cleanest comparison because both validity classes remain represented in both subsets.

\begin{table}[t]
\caption{Llama-3.2-3B validity-probe AUROC conditioned on behavioral
correctness.}
\label{tab:appendix_correct_incorrect}
\centering
\small
\setlength{\tabcolsep}{5pt}
\begin{tabular}{lcccc}
\toprule
\textbf{Split}
& \multicolumn{2}{c}{\textbf{Correct}}
& \multicolumn{2}{c}{\textbf{Incorrect}} \\
\cmidrule(lr){2-3}
\cmidrule(lr){4-5}
& $n$ & \textbf{AUROC}
& $n$ & \textbf{AUROC} \\
\midrule
Random
& 75 & 1.000
& 85 & 1.000 \\

Template
& 98 & 0.970
& 102 & 0.995 \\

Domain
& 71 & 0.989
& 89 & 1.000 \\

Family
& 68 & 1.000
& 92 & 0.994 \\
\bottomrule
\end{tabular}
\end{table}

Pythia-2.8B provides an additional comparison where both gold classes remain available. Under domain holdout, AUROC is $0.965$ on behaviorally correct examples and $1.000$ on incorrectly answered examples. For models whose predictions collapse almost entirely to one answer label, some correctness-conditioned subsets contain only one gold class. AUROC is mathematically undefined in these cases, so these conditions are omitted rather than assigned an artificial score.

\section{Lexical, Metadata, and Shuffled-Label Controls}
\label{app:controls}

We compare hidden-state probes with TF--IDF classifiers using the full prompt, claim only, and premises only. We additionally fit metadata-only classifiers using inference family, domain, difficulty, template-pair identity, prompt-template identity, distractor status, and prompt length.

\begin{table}[t]
\caption{Lexical and metadata control AUROC. Metadata entries give the range
across model-specific evaluations.}
\label{tab:appendix_controls}
\centering
\small
\setlength{\tabcolsep}{4pt}
\begin{tabular}{lcccc}
\toprule
\textbf{Control}
& \textbf{Random}
& \textbf{Template}
& \textbf{Domain}
& \textbf{Family} \\
\midrule
Full prompt
& 0.970
& 0.855
& 0.814
& 0.884 \\

Claim only
& 0.965
& 0.682
& 0.767
& 0.752 \\

Premises only
& 0.500
& 0.500
& 0.500
& 0.500 \\

Metadata only
& 0.498--0.503
& 0.501--0.504
& 0.490--0.504
& 0.501--0.508 \\
\bottomrule
\end{tabular}
\end{table}

The random split contains substantial lexical predictability: both full-prompt and claim-only TF--IDF classifiers perform well in-distribution. These controls therefore make clear that random-split probe performance alone is insufficient for distinguishing broadly transferable validity information from dataset regularities.The premises-only classifier remains at chance, consistent with the matched-pair construction in which valid and invalid examples within a pair share their premise context. Metadata-only prediction also remains close to chance across the primary splits.

For the shuffled-label control, probe training is repeated for 200 random label permutations. Null AUROC means range from $0.496$ to $0.499$, with standard deviations of approximately $0.031$--$0.039$. For every model, the observed random-split AUROC exceeds all 200 shuffled-label results:
\[
p=\frac{1}{200+1}\approx0.005.
\]

Together, these controls motivate placing greater weight on held-out generalization, matched-pair comparisons, and correctness-conditioned evaluation than on the near-perfect random-split AUROC alone.

\section{Secondary Analyses}
\label{app:secondary}

\subsection{Within-Pythia Scaling}
\label{app:scaling}

The two Pythia models provide a limited within-family comparison of parameter scale. Increasing model size from 1.4B to 2.8B improves some forms of transfer but not others.

\begin{table}[t]
\caption{Within-family comparison between Pythia-1.4B and Pythia-2.8B.
Values are validity-probe AUROC.}
\label{tab:appendix_scaling}
\centering
\small
\setlength{\tabcolsep}{5pt}
\begin{tabular}{lcc}
\toprule
\textbf{Evaluation}
& \textbf{Pythia-1.4B}
& \textbf{Pythia-2.8B} \\
\midrule
Random
& 1.000 & 1.000 \\

Template held out
& 0.999 & 0.963 \\

Domain held out
& 0.968 & 0.989 \\

Family held out
& 0.950 & 0.976 \\

Mean domain LOO
& 0.866 & 0.890 \\

Mean family LOO
& 0.859 & 0.866 \\
\bottomrule
\end{tabular}
\end{table}

Domain-held-out AUROC increases from $0.968$ to $0.989$, while family-held-out AUROC increases from $0.950$ to $0.976$. Mean leave-one-domain-out performance also rises from $0.866$ to $0.890$, while mean leave-one-family-out performance changes only modestly from $0.859$ to $0.866$.

Template transfer moves in the opposite direction, decreasing from $0.999$ to $0.963$. This limited two-model comparison therefore does not support a simple monotonic relationship between parameter count and validity decodability.

\subsection{Nominal Difficulty}
\label{app:difficulty}

Difficulty does not produce a consistent behavioral gradient across models. Pythia-1.4B, SmolLM3-3B, and Mistral-7B remain at $0.500$ behavioral accuracy across all three difficulty levels. Pythia-2.8B varies only between $0.498$ and $0.505$. Llama-3.2-3B shows more variation, but not in the expected direction: accuracy increases from $0.435$ on direct one-step examples to $0.475$ on two-step examples and $0.495$ on two-step examples with distractors. Thus, the nominal construction difficulty does not produce a monotonic decline in behavioral accuracy. We therefore treat it as a secondary descriptive analysis rather than as an ordered measure of empirical reasoning difficulty.

\section{Causal Intervention Details}
\label{app:causal}

The intervention uses the layer selected by the random-split probe. Because the probe is fitted on standardized features, its coefficient vector is first mapped back into raw activation coordinates. Let $w_{\ell^*}$ denote the learned standardized probe coefficient vector and $s_{\ell^*}$ the element-wise training-set standard deviation. We define
\[
\tilde v
=
w_{\ell^*}\oslash s_{\ell^*},
\qquad
v
=
\frac{\tilde v}{\|\tilde v\|_2}.
\]

The intervention scale is defined using the training-set standard deviation of projection onto $v$:
\[
\sigma_v
=
\operatorname{SD}_{i\in\mathrm{train}}
\left(h_{i,\ell^*}^{\top}v\right).
\]

We modify the final prompt-token hidden state according to
\[
h'_{i,\ell^*}
=
h_{i,\ell^*}
+
\alpha\sigma_v v,
\]
for
\[
\alpha\in\{-4,-2,-1,0,1,2,4\}.
\]

The primary behavioral outcome is
\[
\Delta M_\alpha(x)
=
M_\alpha(x)-M_0(x),
\]
together with changes in the model's binary verification decision.

\subsection{Random-Direction Controls}

For each model, the probe-derived direction is compared with five random directions orthogonal to $v$. Random interventions are norm matched to the probe-direction perturbation at each intervention strength. This comparison tests whether an observed change is specific to the probe-derived direction rather than a generic consequence of perturbing the hidden state.

\subsection{Full Intervention Sweep}

\begin{figure*}[t]
\centering
\includegraphics[width=0.90\linewidth]{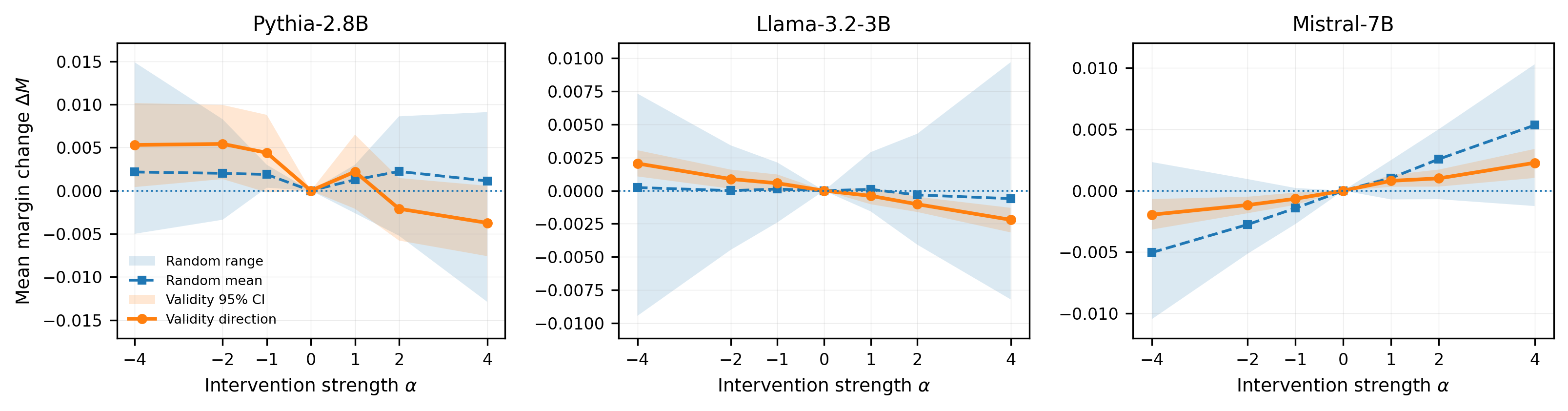}
\caption{Effect of intervention strength on verification output for Pythia-2.8B, Llama-3.2, and Mistral-7B. The full sweep uses $\alpha\in\{-4,-2,-1,0,1,2,4\}$. Changes along the probe-derived validity direction remain small and are not consistently larger than those produced by norm-matched random orthogonal directions.}
\label{fig:causal_dose_response}
\end{figure*}

Across the intervention sweep, changes in binary predictions are rare. No prediction flips occur for any model at $\alpha=+4$. At $\alpha=-4$, only Llama-3.2 changes a binary decision, with one flip among 160 evaluated examples ($0.625\%$).

\subsection{Matched-Projection Patching}

We additionally use the naturally occurring probe-direction difference between the valid and invalid member of a matched pair. For target state $h_i$ and matched source state $h_j$, define 
\[
\delta_{j\rightarrow i}
=
v^\top h_j-v^\top h_i.
\]

We then replace only the target's projection along $v$:
\[
h'_i
=
h_i+\delta_{j\rightarrow i}v.
\]

Mean output-margin changes remain small: approximately $0.0020$--$0.0045$ for Pythia-2.8B, at most $0.0011$ in magnitude for Llama-3.2-3B, and approximately $0.0011$--$0.0012$ for Mistral-7B. These interventions test a deliberately narrow causal hypothesis. Their weak effect does not establish that validity-related information is causally irrelevant to the model's computation. Rather, the results do not support the probe-derived linear direction as a sufficient low-dimensional control variable at the tested intervention site.
\end{document}